\documentclass{article}

\usepackage[utf8]{inputenc}
\usepackage[T1]{fontenc}

\usepackage[margin=1in]{geometry}
\usepackage{authblk}  

\usepackage{amsmath}
\usepackage{amsfonts}
\usepackage{amssymb}

\usepackage{graphicx}
\usepackage{booktabs}
\usepackage{multirow}

\usepackage{listings}

\usepackage{microtype}
\usepackage{nicefrac}

\usepackage{xcolor}
\usepackage{url}
\usepackage{hyperref}

\usepackage[numbers,compress]{natbib}

\definecolor{lightgray}{gray}{0.95}

\hypersetup{
    colorlinks=true,
    linkcolor=blue,
    citecolor=blue,
    urlcolor=blue
}

\usepackage{booktabs} 
\usepackage{xcolor}   

\definecolor{best}{HTML}{0056B3} 
\definecolor{bad}{HTML}{D97706}  

\newcommand{\bestval}[1]{\textcolor{best}{\textbf{#1}}}

\usepackage{adjustbox}
\usepackage{placeins}

\title{Sheet As Token: A Graph-Enhanced Representation\\
for Multi-Sheet Spreadsheet Understanding}

\author{
  Sheet Agent Research Team
}
\date{}

\begin{document}

\maketitle


\begin{abstract}
Workbook-scale spreadsheet understanding is increasingly important for language-model-based data analysis agents, but remains challenging because relevant information is often distributed across multiple sheets with heterogeneous schemas, layouts, and implicit relationships. 
Existing retrieval-augmented approaches typically decompose spreadsheets into rows, columns, or blocks to improve scalability; 
however, such chunk-centric representations can fragment worksheets into isolated text spans and weaken global sheet-level semantics. 
We propose \textsc{Sheet As Token} (SAT), a graph-enhanced framework that treats each worksheet as a unified semantic unit for multi-sheet spreadsheet retrieval.
SAT serializes sparse schema-aware features, including sheet name, shape, and
column headers, and encodes each worksheet into a compact dense token.
Given a query, SAT retrieves candidates with a BGE-initialized Sheet Encoder
and refines them with a gated relational GNN.
In strict full-corpus evaluation, SAT reaches 0.9173 NDCG@5 on
IndustryTab-614 and 0.9222 on IndustryTab-1K, relative improvements of
44.6\% and 46.7\% over zero-shot BGE RAG, respectively.
SAT therefore improves both retrieval accuracy and serving efficiency:
on IndustryTab-1K, it exceeds a Qwen3.5-9B RAG reranker by 12.5\% while
reducing online latency from 2.61\,s to 9.24\,ms, approximately
$283\times$ faster.
These results show that SAT provides accurate and latency-efficient retrieval
in the evaluated fixed-corpus setting\footnote{Code and data are available at
\url{https://github.com/SHITIANYU-hue/SheetasToken}}.
\end{abstract}

\section{Introduction}
\label{sec:intro}

Spreadsheets remain one of the most widely used formats for organizing, analyzing, and exchanging structured information in enterprise and analytical workflows. They are used for financial reporting, project tracking, sales analysis, inventory management, human-resource records, and many other operational tasks. Recent work has shown that spreadsheet understanding is an important and challenging problem for large language models, because spreadsheets differ substantially from plain text documents: they encode information through worksheets, headers, cell values, formulas, layouts, and workbook-level organization ~\cite{dong2024spreadsheetllm, li2023sheetcopilot}. As a result, many real-world spreadsheet tasks require not only understanding an individual table, but also identifying how multiple sheets within a workbook jointly support a user request.

In this paper, we study the problem of multi-sheet spreadsheet understanding through query-conditioned sheet retrieval. Given a natural language query and a collection of workbook sheets, the goal is to retrieve the set of sheets that jointly provide the evidence needed for downstream reasoning, question answering, or code generation. This task requires two complementary capabilities. First, the system must understand the semantic role of each sheet, including what the sheet represents, what attributes it contains, and how it is structurally organized. Second, it must model dependencies among sheets, since the answer to a query may depend on multiple related worksheets rather than a single isolated table.

A straightforward solution is to serialize the entire workbook and feed it into a long-context language model. However, this approach is expensive, noisy, and difficult to scale. Enterprise workbooks may contain many sheets, large grids, heterogeneous cell types, repeated headers, irrelevant rows, and redundant local values. Full-workbook serialization forces the model to process substantial irrelevant content before identifying the supporting evidence. It also makes the retrieval decision dependent on the model's ability to sift through a long, semi-structured context, which becomes increasingly inefficient as workbook size grows.

Retrieval-augmented generation provides a natural way to reduce this burden by retrieving only relevant external evidence before generation ~\cite{lewis2020rag}. Dense retrieval methods further show that textual units can be embedded into a shared semantic space and retrieved efficiently for downstream question answering ~\cite{karpukhin2020dpr}. Following this direction, recent spreadsheet-oriented retrieval systems decompose workbooks into fine-grained units such as rows, columns, blocks, or textual chunks, and then retrieve these units using lexical, dense, or hybrid retrieval strategies ~\cite{gulati2026frtr}. Such decomposition improves scalability because it avoids encoding the entire workbook at once.

However, fine-grained decomposition also changes the retrieval primitive in a way that may be misaligned with multi-sheet spreadsheet understanding. A worksheet is not merely a collection of independent rows, columns, or cell spans. It is a functional unit with a name, schema, shape, representative values, and relationships to other worksheets. Once a workbook is broken into isolated chunks, global sheet identity and cross-sheet dependencies may be weakened or lost. Consequently, a system may retrieve locally similar fragments while missing the coherent set of sheets needed for reasoning across a workbook. This limitation is especially important for queries whose evidence is distributed across temporally related sheets, structurally aligned worksheets, or functionally complementary tabs.

\begin{figure*}[ht]
\centering
\begin{minipage}[c]{0.30\textwidth}
    \centering
    \includegraphics[width=\linewidth]{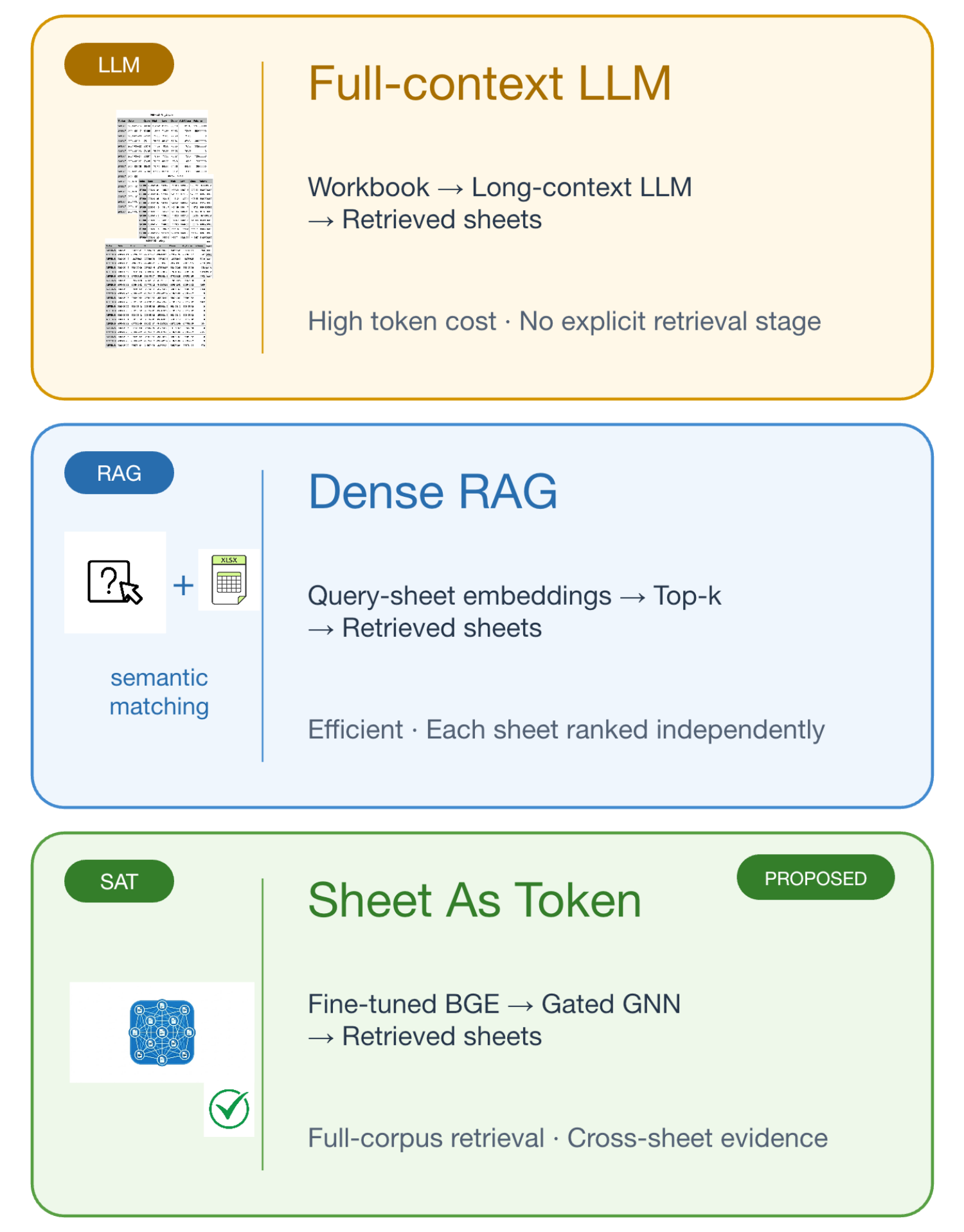}
\end{minipage}\hfill
\begin{minipage}[c]{0.68\textwidth}
    \centering
    \includegraphics[width=\linewidth]{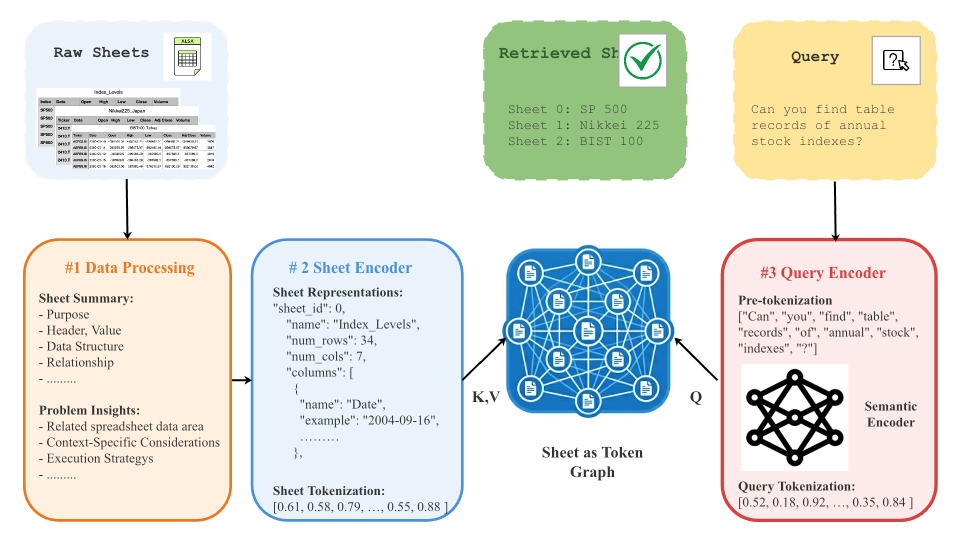}
\end{minipage}
\caption{Comparison and overview of the proposed \textsc{Sheet As Token} (SAT)
framework. \textbf{Left:} full-context LLM selection, dense RAG, and SAT use
different retrieval primitives. \textbf{Right:} SAT converts raw workbook
sheets into schema-aware records, retrieves full-corpus candidates with the
fine-tuned BGE Sheet Encoder, and applies query-conditioned graph reranking to
return the supporting sheets.}
\label{fig:overview}
\end{figure*}

This raises a natural question: if multi-sheet understanding ultimately requires identifying relevant worksheets, should the sheet itself be treated as the basic semantic unit for retrieval? Instead of starting from rows, columns, or blocks, we revisit the representation level of spreadsheet retrieval and argue that each sheet should be represented as a unified object. This perspective is analogous to treating a worksheet as a high-level semantic token in the workbook, preserving its global role while avoiding the cost of full-grid encoding.

Based on this idea, we propose \textsc{Sheet As Token} (SAT), a unified
sheet-level framework for multi-sheet spreadsheet understanding. Rather than
encoding raw grids or ingesting full rows, SAT serializes the sheet name,
column headers, and shape into a structured semantic record and encodes it
into a dense vector, which we call a \emph{Sheet Token}. The goal is not to
preserve every cell value, but to retain the essential semantic and structural
signals needed to distinguish worksheets and support sheet-level retrieval.

Nevertheless, isolated sheet tokens are insufficient for multi-sheet reasoning. Supporting evidence is often relational: a relevant sheet may be connected to other sheets through schema overlap, temporal alignment, shape compatibility, semantic similarity, or query-conditioned co-relevance, even when it does not share direct lexical overlap with the query. Graph neural networks provide a natural inductive bias for modeling such relational structure, since they propagate information through dependencies among connected entities ~\cite{battaglia2018relational, zhou2020gnn}. At the same time, spreadsheets differ from formal relational databases: they often lack explicit schemas, violate normalization assumptions, and encode important relationships implicitly through names, headers, layouts, and workbook organization, unlike the more explicit relational structures studied in recent relational learning benchmarks ~\cite{robinson2024relbench}.

To address this challenge, SAT includes a graph-enhanced relational retriever
for cross-sheet reasoning. As shown in \autoref{fig:overview}, given a query
and Stage~1 candidates, it constructs a query-specific graph whose nodes are
sheets and whose edges encode semantic, structural, and explicit spreadsheet
dependencies. A gated residual GNN learns small relational corrections to a
strong listwise base ranking, rather than replacing the Stage~1 ordering with
unconstrained message passing.

SAT is trained as a deployment-matched, two-stage full-corpus cascade. A
BGE-initialized \textbf{Sheet Encoder} first learns reusable Sheet Tokens with
query--sheet contrastive supervision; a relational \textbf{Graph Retriever}
then reranks the actual candidates returned by Stage~1. We evaluate this
pipeline on IndustryTab-614 and IndustryTab-1K, comprising 614 and 1,002
sheets and 1,453 and 1,797 queries, respectively. SAT achieves 0.9173 and
0.9222 NDCG@5 while keeping the online path lightweight: sheet representations
are cached offline, and retrieval plus graph reranking takes only 9.24\,ms per
query on an A40. The resulting system therefore improves retrieval quality
without incurring the latency of online LLM reranking.

In summary, this paper makes the following contributions:

\begin{itemize}
    \item We introduce \textsc{Sheet As Token} (SAT), a sheet-level retrieval
    abstraction that encodes sparse schema-aware metadata into reusable,
    cacheable tokens, preserving worksheet identity without processing full
    grids or fragmenting sheets into chunks.
    \item We develop a gated residual graph retriever that learns relational
    corrections over real full-corpus candidates, combining semantic,
    structural, and explicit spreadsheet dependencies while preserving the
    strong Stage~1 ranking.
    \item We establish strict full-corpus evaluation at two scales. SAT reaches
    0.9173 and 0.9222 NDCG@5, improving over zero-shot BGE RAG by 44.6\% and
    46.7\%. On IndustryTab-1K, SAT outperforms a Qwen3.5-9B RAG reranker by
    12.5\% and is approximately $283\times$ faster online (9.24\,ms versus
    2.61\,s).
\end{itemize}
\section{Related Work}
\label{sec:related}

\subsection{Table Understanding Datasets and Benchmarks}

\paragraph{Procedural schema-matching fabricators.}
The Valentine experiment suite and its dataset generator
\cite{koutras2021valentine,valentinefabricator} produce paired
\texttt{source} and \texttt{target} tables from a single input table via
controlled horizontal/vertical partitioning and noise injection.
While this yields deterministic column-to-column ground truth for evaluating pairwise semantic alignment, it fails to capture inter-sheet dependencies or workbook-scale context.
Consequently, we adopt Valentine's fabricator to provide column-level supervision within our broader pipeline.

\paragraph{QA and retrieval over real tables.}
A second line of work curates QA and retrieval tasks over real-world tables to evaluate long-context reasoning~\cite{longtablebench2025} and realistic query distributions.
While these benchmarks cover diverse intents and scales, from single-table
reasoning suites~\cite{chen2019tabfact, zhu2021tatqa} to spreadsheet-scale
understanding, question answering, and manipulation
\cite{li2023sheetcopilot, dong2024spreadsheetllm, wang2025sheetbrain}, they do
not jointly provide the column-level alignment supervision and explicit
inter-sheet topology required by our retrieval setting.
Because no existing dataset provides both forms of supervision simultaneously, we construct a novel corpus in Section~\ref{sec:data_preparation} to train and evaluate the two-stage method described in Section~\ref{sec:method}.

\subsection{Retrieval-Augmented Generation for Structured Data}

Retrieval-augmented generation (RAG)~\cite{lewis2020rag} has been increasingly adapted to tabular settings to bypass the scalability bottlenecks inherent in full-context encoding.
Unlike traditional open-domain pipelines that process documents as flat, linear text sequences, spreadsheet-based retrieval must actively model the multi-dimensional, semi-structured topology of the data~\cite{glass2021capturing}.
Current state-of-the-art approaches, such as FRTR~\cite{gulati2026frtr}, attempt to maintain scalability by decomposing Excel workbooks into granular row, column, and block embeddings, which are then surfaced via hybrid lexical-dense retrieval using Reciprocal Rank Fusion (RRF).

However, chunk-based decomposition often reduces spreadsheet components to isolated text spans, which may weaken global sheet semantics and cross-sheet dependencies. Drawing inspiration from contrastive representation learning and dual-encoder retrieval architectures~\cite{karpukhin2020dpr}, we instead treat each individual sheet as a unified, cohesive entity within a latent semantic space. The resulting sheet-level representation is designed to prioritize structural and functional consistency over localized keyword matching. Combined with query-conditioned graph reasoning, this design is intended to improve retrieval when relevant sheets share structural, functional, or relational signals even with limited direct lexical overlap.

\subsection{Graph Neural Networks for Relational Data}

Graph Neural Networks (GNNs) provide a powerful framework for relational data by embedding structural inductive biases into iterative message-passing architectures~\cite{battaglia2018relational, zhou2020gnn}.
Unlike $k$-nearest neighbor ($k$-NN) methods that rely on static, feature-space proximity~\cite{cover1967nearest, hamilton2020graph}, GNNs explicitly model complex, heterogeneous interactions through trainable edge aggregation, capturing dependencies that extend beyond simple geometric similarity.
While recent frameworks like RelBench~\cite{robinson2024relbench} successfully apply GNNs to formal relational databases, enterprise spreadsheets present a fundamentally different challenge: they lack explicit schemata, heavily violate normalization principles, and encode critical dependencies implicitly through spatial layouts and formulas.
To address this gap, our Graph-Enhanced Cross-Sheet Reasoning module specifically adapts graph-based representation learning for this highly semi-structured, schema-free environment.

\section{Methodology}
\label{sec:method}

\subsection{Problem Formulation}

Let a workbook collection $\mathcal{W}$ consist of $n$ sheets, denoted as $\mathcal{W}=\{S_{1}, S_{2}, \dots, S_{n}\}$.
Each sheet $S_{i}$ is defined as a two-dimensional grid of cells of size $R_{i} \times C_{i}$, where $R_{i}$ and $C_{i}$ represent the number of rows and columns, respectively.
Each cell $c_{r,c}^{(i)}$ within $S_{i}$ can contain heterogeneous data modalities, including numerical values, unstructured text, or explicit formulaic dependencies.

Our primary objective is to map this discrete, semi-structured workbook into a continuous latent semantic space.
Specifically, we aim to learn an encoder that represents each sheet $S_i$ as
a single dense vector $\mathbf{z}_i \in \mathbb{R}^d$, which we define as the
\textit{Sheet Token} used throughout SAT.
This transformation effectively encodes the entire workbook $\mathcal{W}$ as a constellation of points in a $d$-dimensional manifold, capturing the macro-architecture of the spreadsheet.

Given a natural language query $q$ and candidate sheets $\{S_i\}_{i=1}^{n}$, the retrieval objective is to predict a supporting sheet set $Y_q \subseteq \{1,\ldots,n\}$. The model assigns each candidate sheet a relevance score
\[
r_i = g_\theta(q, S_i, \mathcal{G}_q),
\]
where $\mathcal{G}_q$ denotes the query-specific candidate graph constructed over the candidate sheets. The predicted support set is obtained by selecting the top-ranked sheets or sheets whose relevance scores exceed a threshold. This formulation focuses on retrieval rather than final answer generation; the retrieved sheet set can be used by downstream QA, reasoning, or code-generation systems.

\subsection{Feature Extractor}
\label{sec:feature_extractor}

\begin{figure}[!ht]
\centering
\includegraphics[width=\linewidth]{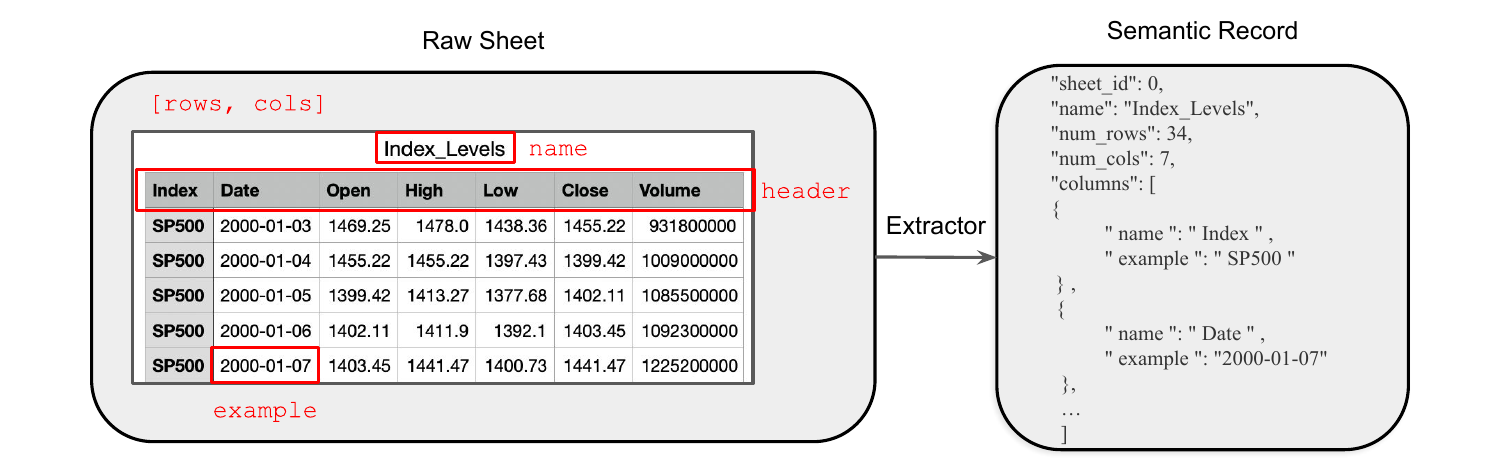}
\caption{Transformation of a raw spreadsheet into a schema-aware semantic record. The feature extractor maps the sheet's name, grid dimensions, column headers, and a value example into a structured JSON format, isolating structural semantics while discarding redundant row-level content.}
\label{fig:feature_extractor}
\end{figure}

Before projecting a sheet $S_i$ into the \textit{Sheet Token} space, we decompose it into four typed feature channels, as illustrated in Figure~\ref{fig:feature_extractor}.
This extends the schema-aware paradigm established by Valentine~\cite{koutras2021valentine}.
Each sheet is materialized as a semantic record containing the following fields:

\paragraph{(i) Sheet name.}
The concatenation of the workbook filename and the tab identifier.
This provides critical high-level domain, entity, and temporal semantics as the strongest single distinguishing signal.

\paragraph{(ii) Column headers.}
The canonical column names, which define the sheet's structural schema.
Explicitly encoding this array (e.g., \texttt{["Index", "Date", \dots]}) provides the model with a lexical map of the data attributes.
This captures the table's semantic axes, which is essential for differentiating sheets that share similar names but contain divergent fields.

\paragraph{(iii) Example value.}
The first non-null cell per column, truncated to 60 characters.
Extracting a concrete instance (e.g., \texttt{2000-01-07}) resolves overloaded headers without requiring full-row ingestion.
High-cardinality text is replaced with typed placeholders to mitigate vocabulary inflation.

\paragraph{(iv) Shape.}
The grid dimensions $(R_i, C_i)$, which map directly to the \texttt{num\_rows} and \texttt{num\_cols}.

\medskip
\noindent
Together, these channels form a sparse, schema-aware representation capturing \emph{what the table is about}, \emph{what it contains}, and \emph{how it is shaped}.
By relying on a unified lookup into this preprocessed state, we ensure all downstream stages are evaluated on consistent, deduplicated inputs.

\subsection{Sheet As Token: Unified Representation via Sparse Feature Extraction}
\label{sec:sat}

\begin{figure}[!ht]
\centering
\includegraphics[width=0.8\linewidth]{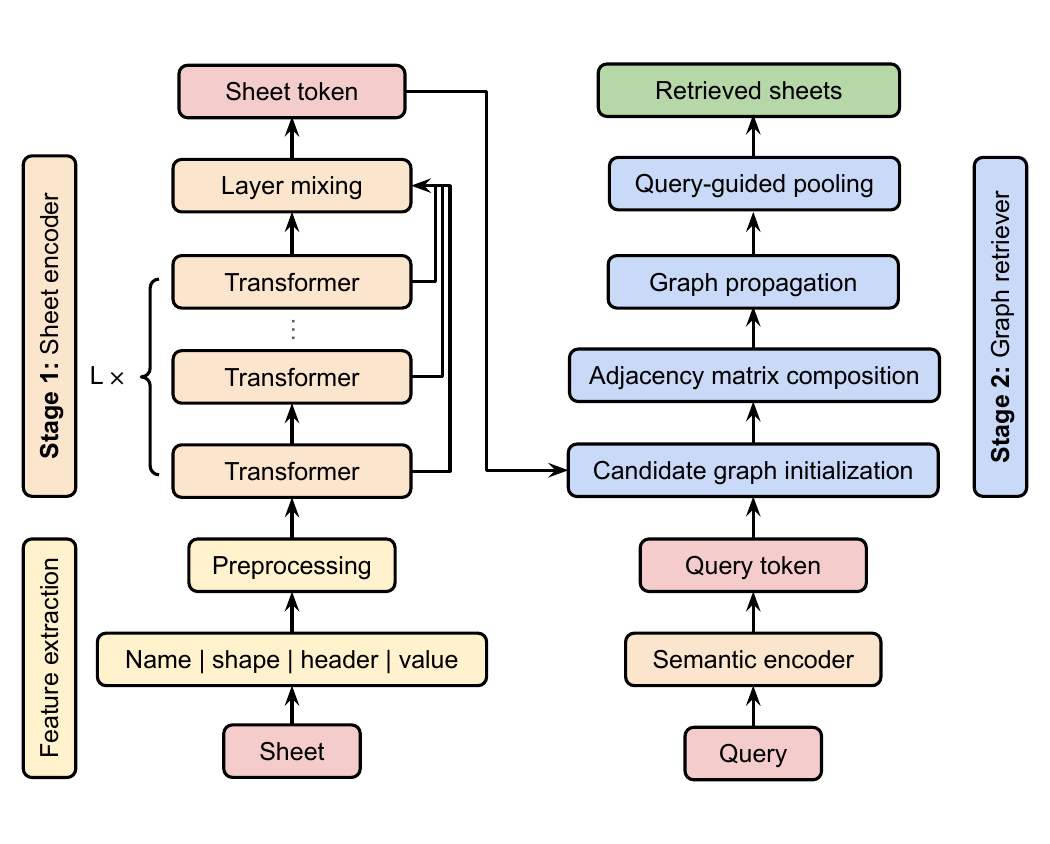}
\caption{Architecture of SAT. \textbf{Stage 1} uses a BGE-based bi-encoder to
map sparse sheet records into dense Sheet Tokens. \textbf{Stage 2} reranks
full-corpus candidates with the proposed gated relational GNN.}
\label{fig:architecture}
\end{figure}

The core of SAT is to map each sheet $S_i$ into a single dense vector
$\mathbf{z}_i \in \mathbb{R}^d$, which we term the \textit{Sheet Token}.
As \autoref{fig:architecture}, our architecture utilizes the structured, schema-aware records defined before to represent the table strictly through its semantic metadata.

\paragraph{Linearized Feature Serialization.}
We first linearize the extracted feature record into a unified textual sequence $X_i$.
The sheet dimensions ($R_i \times C_i$), sheet name ($s_i$), and up to $k=12$ canonical column headers ($h_j$) are concatenated using a deterministic structural template:
\begin{equation}
    X_i = [\text{source:\ } s_i] \oplus [\text{shape:\ } R_i \times C_i] \oplus \left[\text{headers:\ } \bigoplus_{j=1}^{k} h_j \right]
\end{equation}
where $\oplus$ denotes string concatenation with appropriate delimiters (e.g., semi-colons and pipes).
This specific serialization preserves the macroscopic identity of the sheet without exposing the model to the computational overhead or noise of full-row ingestion.

\paragraph{BGE backbone and pooling.}
The serialized sequence $X_i$ is tokenized and processed by
\texttt{BAAI/bge-base-en-v1.5}, truncated to a maximum context length of 512
tokens. We use the same bi-encoder to embed the natural-language query and the
sheet record.
Let $\mathbf{H}_i \in \mathbb{R}^{T \times d_{\text{hidden}}}$ denote the
final hidden states for the sequence of length $T$. The pooled and
$\ell_2$-normalized representation is the Sheet Token:
\begin{equation}
    \mathbf{z}_i =
    \operatorname{norm}\!\left(\operatorname{Pool}(\mathbf{H}_i)\right)
    \in \mathbb{R}^d .
\end{equation}

\paragraph{Objective Function.}
Stage~1 is optimized for the same query--sheet retrieval task used at
inference. For a query $q$, its annotated relevant sheets form the positive
set $\mathcal{P}_q$; in-batch sheets and BGE-retrieved non-relevant sheets form
the negative set $\mathcal{N}_q$. We minimize a multi-positive contrastive
loss
\begin{equation}
 \mathcal{L}_{\mathrm{ret}} =
 -\log
 \frac{\sum_{p\in\mathcal{P}_q}
 \exp(\operatorname{sim}(\mathbf{z}_q,\mathbf{z}_p)/\tau)}
 {\sum_{s\in\mathcal{P}_q\cup\mathcal{N}_q}
 \exp(\operatorname{sim}(\mathbf{z}_q,\mathbf{z}_s)/\tau)} ,
\end{equation}
where $\tau$ is the temperature. We fine-tune the final four BGE layers and
cache all sheet tokens after training. At inference, cosine similarity over
the entire sheet corpus supplies the top-$K$ candidates to Stage~2.

\subsection{Graph-Enhanced Cross-Sheet Reasoning}
\label{sec:gcsr}

While sheet-level representations capture the local semantics of individual tables, multi-sheet spreadsheet understanding requires reasoning over interconnected \emph{sets} of sheets rather than isolated tables.
For a given query, supporting evidence often spans multiple semantically related, structurally aligned, or shape-compatible sheets.
To model these dependencies, SAT Stage~2 constructs a query-specific candidate
graph and performs cross-sheet reasoning before final retrieval.

\paragraph{Candidate graph construction.}
Given a query $q$, we form a candidate workspace $\mathcal{C}(q)=\{S_1,\dots,S_m\}$, where each node corresponds to a candidate sheet.
The query and candidate sheets are mapped into a shared embedding space using the Stage 1 Sheet Encoder.
We construct multiple adjacency priors to capture complementary cross-sheet relations.
In our setting, these priors encode four signals: semantic similarity, query-conditioned relevance, schema consistency, and shape compatibility.
Together, these channels provide a structured relational view of the candidate workspace under the current query.

\paragraph{Graph retriever.}
A listwise base scorer first produces a relevance logit for every retrieved
candidate. The relational GNN then aggregates messages separately over the
active adjacency channels and applies a learned residual gate:
\begin{equation}
 \mathbf{h}^{(\ell+1)}_i =
 \operatorname{LN}\!\left(
 \mathbf{h}^{(\ell)}_i +
 \sigma(g_\ell)\,
 \Delta^{(\ell)}_i
 \right).
\end{equation}
The gate is initialized near zero so that Stage~2 begins from the selected
base ranking and learns only evidence-supported graph corrections. Formula,
summary-source, aggregation, and join channels can be enabled independently;
validation selects the active subset for each corpus.

\paragraph{Training objective.}
The model is trained on query-centered candidate sets containing both positive and negative sheets.
We optimize the retriever using listwise softmax supervision, node-level
binary relevance supervision, and a pairwise margin loss between annotated
relevant sheets and retrieved hard negatives.
This design jointly supervises workspace-level retrieval and fine-grained node discrimination, enabling the model to learn both which sheet set supports the query and which individual sheets are responsible for that support.

\paragraph{Summary.}
Overall, SAT's two-stage design provides a principled mechanism for
query-conditioned cross-sheet retrieval.
Detailed formulations are provided in Appendix~\ref{app:graph_details}.%

\section{Experiments}
\label{sec:exp}

\subsection{Data preparation}
\label{sec:data_preparation}

We evaluate two related, separately indexed corpora: IndustryTab-614
(\textbf{small dataset}) and IndustryTab-1K (\textbf{large dataset}). The
suffixes indicate corpus scale; the corpora do not share a
\texttt{sheet\_id} namespace. Each is serialized into three JSON files:
\begin{itemize}\itemsep=2pt\topsep=2pt\parskip=0pt
\item \texttt{sheets.json} maps each \texttt{sheet\_id} to a record of (\textsc{filename}, \textsc{dimensions}, $\{\textsc{header},\ \textsc{example}\}$).
\item \texttt{query.json} associates each natural-language query with disjoint
positive and negative sheet-ID lists. The current releases of both corpora
also store hard-negative IDs, query and dependency types, and whether a query
is dependency-heavy.
\item \texttt{train.json} contains historical same-source sheet pairs. We
retain it in the release for provenance but do not use it in the reported
full-corpus query--sheet experiments.
\end{itemize}

\paragraph{Label preparation.}
Each query's positive set is defined by sheets sharing a key attribute (company, fiscal year, quarter, sub-category, or two-attribute combinations).
An equal-sized negative set is sampled uniformly from the remaining sheets.
For both corpora, we further annotate schema-matched but entity-mismatched
hard negatives and queries requiring temporal/schema alignment, join-key
reasoning, aggregation, formula tracing, or summary-source tracing.

\paragraph{Dataset statistics.}
\autoref{tab:dataset_statistics} summarizes the two corpora.
IndustryTab-614 contains ten spreadsheet categories and supports the original
representation-learning and graph-retrieval experiments. Its current
1,453-query release comprises 583 semantic lookups, 252 comparisons, 248
aggregations, 220 join-key queries, 97 formula-tracing queries, and 53
summary-source queries; 983 are dependency-heavy. IndustryTab-1K expands the
temporal and entity coverage and adds Budget Planning. Its 1,797 queries comprise 631
semantic lookups, 357 comparisons, 339 aggregations, 257 join-key queries, 140
formula-tracing queries, and 73 summary-source queries; 1,319 are marked as
dependency-heavy. Detailed category counts and dependency statistics are
reported in Appendix~\ref{app:data}.

The original arXiv snapshot contained 134 listwise queries over
IndustryTab-614. We retain that file only for provenance and do not use it in
the full-corpus experiments: all new small-corpus results use the server
release with 1,453 queries, richer dependency labels, and hard negatives.

\paragraph{Relationship between the corpora.}
IndustryTab-1K is an identity-level expansion rather than an independent-domain
test set: all 614 original sheet names reappear in the 1,002-sheet pool. Of
these, 590 retain identical extracted metadata and 24 have updated metadata;
the expanded pool adds 388 new sheet identities. The query suites are largely
distinct (only one exact query string is shared). We therefore use the second
corpus to test scale, harder negatives, and richer dependency supervision, not
to claim zero-overlap cross-dataset generalization.

\begin{table*}[t]
\centering
\small
\caption{Statistics of the two full-corpus evaluation benchmarks.
``Avg.\ pos./neg.'' counts query-level annotations before retrieval.}
\label{tab:dataset_statistics}
\begin{tabular}{lrrrl}
\toprule
\textbf{Dataset} & \textbf{Sheets} & \textbf{Queries} &
\textbf{Avg.\ pos./neg.} &
\textbf{Additional annotations} \\
\midrule
Small: IndustryTab-614 & 614 & 1,453 & 8.06 / 7.86 &
hard negatives, query/dependency types \\
Large: IndustryTab-1K  & 1,002 & 1,797 & 8.05 / 8.05 &
hard negatives, query/dependency types \\
\bottomrule
\end{tabular}
\end{table*}

\begin{table*}[t]
\centering
\small
\setlength{\tabcolsep}{5pt}
\caption{Compared methods in increasing order of adaptation and structured
reasoning. RAG and all SAT variants share the same Feature Extractor
serialization.}
\label{tab:compared_methods}
\begin{tabular}{lcccc}
\toprule
\textbf{Method} & \textbf{BGE adapted} & \textbf{Stage 2}
& \textbf{Graph reasoning} & \textbf{Candidate/context path} \\
\midrule
LLM-only & -- & Local LLM & -- & all sheet names $\rightarrow 5$ \\
RAG & No & -- & -- & full corpus $\rightarrow 5$ \\
RAG+LLM & No & Local LLM & -- & full corpus $\rightarrow 50 \rightarrow 5$ \\
\midrule
SAT (Stage 1 only) & Yes & -- & -- & full corpus $\rightarrow 5$ \\
SAT (Cross-Encoder) & Yes & Cross-Encoder & -- &
full corpus $\rightarrow 50 \rightarrow 5$ \\
SAT (ours) & Yes & Gated GNN & Yes &
full corpus $\rightarrow 50 \rightarrow 5$ \\
\bottomrule
\end{tabular}
\end{table*}

\paragraph{Dataset split.}
IndustryTab-614 uses an $80/20$ query split, followed by a 90/10
train--validation split within the training portion: 1,047 training, 116
validation, and 290 test queries. IndustryTab-1K uses the fixed seed-42
$90/10$ query split, followed by the same validation construction: 1,457
training, 161 validation, and 179 test queries. Model selection uses
validation NDCG@5 only. The small-corpus results average
independently shuffled 80/20 splits for seeds 42--44; the large-corpus results
hold the seed-42 split fixed and vary only training seeds.

\subsection{Experimental Protocol}
\label{sec:experiment_setup}

\paragraph{Controlled input serialization.}
The Feature Extractor stores all four channels defined in
\autoref{sec:feature_extractor}, but the full-corpus retrieval experiments use
the same controlled serialization of sheet name, shape, and up to 12 headers
for RAG and every SAT variant. Example values are therefore not used
in the reported full-corpus comparison.

\paragraph{Training forward flow.}
\textbf{SAT Stage 1} initializes the \textbf{Sheet
Encoder} from BGE and learns reusable sheet-level representations from
query--positive-sheet contrastive supervision.
The forward pass fetches sheet metadata, serializes each sheet into a
structured textual record, and encodes queries and sheets with a shared
bi-encoder backbone.
Retrieved non-relevant sheets provide hard negatives.

In \textbf{SAT Stage 2}, the \textbf{Graph Retriever} performs
query-conditioned listwise reranking over real Stage 1 top-50 candidates.
The query and candidate sheets are encoded using the frozen Sheet Encoder
from Stage 1. The retriever constructs a multi-channel adjacency prior from
semantic, query-conditioned, schema, shape, and explicit dependency relations.
Gated residual message passing refines a listwise base scorer while preserving
its ordering when graph evidence is unhelpful.
The retriever is optimized with listwise, node-level binary, and relevant
versus hard-negative pairwise losses.

\paragraph{Inference execution.}
During inference, the Sheet Encoder provides semantic representations for candidate sheets, while the Stage 2 Graph Retriever performs query-specific cross-sheet reasoning over the candidate workspace.
Given a natural-language query, the system encodes the query and candidate sheets, dynamically constructs a query-conditioned interaction graph, performs cross-sheet message passing, and outputs final relevance scores for the candidate set.
After training, sheet representations can be precomputed or cached for efficient retrieval, while final ranking decisions remain grounded in query-specific inter-sheet relationships rather than isolated query-sheet similarity.

\paragraph{Evaluation metrics.}
For strict full-corpus retrieval on both datasets, we report NDCG@5, macro
multi-label Recall@5, Hit@5, MRR@5, and HN-FPR@1. Macro Recall@5 averages,
over queries, the fraction of all annotated relevant sheets retrieved in the
top five; Hit@5 measures whether at least one relevant sheet is retrieved.
HN-FPR@1 is the fraction of eligible queries (those with an annotated hard
negative) whose top-ranked prediction is a hard negative.

\subsection{Compared Methods}
\label{sec:compared_methods}

We organize the comparison as a progression from training-free retrieval to
the complete proposed system. Except for LLM-only, every method consumes the
same controlled output of the Feature Extractor in
\autoref{sec:feature_extractor}: sheet name, shape, and up to 12 headers.
Thus, differences among RAG and SAT variants measure adaptation and reranking
rather than access to richer sheet content.

\paragraph{Training-free baselines.}
\textbf{LLM-only} gives a local Qwen3.5-9B Q4 model the names of all
sheets and asks it to return five IDs; this compact input already occupies
24{,}603 tokens per query. \textbf{RAG} embeds the common feature
serialization with zero-shot BGE-base-en-v1.5 and directly returns its top
five sheets. \textbf{RAG+LLM} first retrieves 50 sheets with the same RAG
retriever, then asks the same local LLM to rerank their common feature
records.

\paragraph{SAT variants.}
\textbf{SAT (Stage 1 only)} fine-tunes the final four BGE layers with
query--positive-sheet contrastive learning and retrieved hard negatives, then
returns the top five sheets without Stage~2. \textbf{SAT (Cross-Encoder)}
independently scores query--sheet pairs in the real Stage~1 top-50 candidate
set. \textbf{SAT} is our complete method: its gated GNN retriever learns
relational corrections over the same candidates using spreadsheet dependency
channels.

\autoref{tab:compared_methods} summarizes the resulting progression from
training-free retrieval to the complete graph-enhanced SAT pipeline.

\subsection{Main Results: Full-Corpus Retrieval}
\label{sec:end_to_end}

We evaluate the complete retrieval path against the \textbf{entire corpus}:
290 queries search all 614 sheets in IndustryTab-614, while 179 queries search
all 1,002 sheets in IndustryTab-1K. No positive sheet is injected into either
candidate set. This experiment evaluates
end-to-end \emph{evidence-sheet retrieval}. The dataset does not contain
reference final answers, so we do not report answer-generation accuracy.

\subsubsection{Training-Free Baselines}

We compare a dense RAG retriever based on BGE-base-en-v1.5 and two local-LLM
settings. The RAG+LLM system retrieves 50 sheets with BGE and asks a local
quantized Qwen3.5-9B (Q4\_K\_M) model to rerank rich sheet records
(name, shape, and up to 12 headers). The direct-LLM baseline instead receives
the names of every sheet in the corpus in one prompt. Both LLM settings run through
Ollama with temperature zero, structured output for exactly five distinct
sheet IDs, and a 32,768-token context window.
RAG and RAG+LLM use exactly the same Feature Extractor serialization as the
SAT variants.
For four of 179 RAG+LLM queries, the model returned only two to four valid
IDs; we retained those choices and deterministically filled the remaining
positions with the original RAG order.

\subsubsection{SAT on Both Datasets}

The preceding baselines expose candidate recall as the dominant bottleneck.
We therefore initialize Stage 1 with
\texttt{BAAI/bge-base-en-v1.5}~\cite{xiao2023cpack}, use the common name,
shape, and header serialization, and fine-tune the final four encoder
layers with query--positive-sheet contrastive learning and retrieved hard
negatives. Stage 1 searches the complete corpus and passes its top 50 sheets
to Stage 2. Our complete method uses a gated relational GNN whose residual
updates start near zero and learn small message-passing corrections.
Validation removes the dense join-key channel on both corpora: the selected
IndustryTab-614 graph uses aggregation, formula-reference, and summary-source
edges, while IndustryTab-1K uses formula-reference and summary-source edges.

\begin{table*}[t]
\centering
\scriptsize
\setlength{\tabcolsep}{2.7pt}
\caption{Strict full-corpus comparison on both datasets. IndustryTab-614
reports mean $\pm$ standard deviation over three 80/20 query splits (seeds
42--44). IndustryTab-1K uses one fixed 90/10 split; SAT results average three
training seeds. Stage~2 uses real top-50 candidates with no positive injection.
Latency is mean online time per query on an NVIDIA A40 with batch size one;
offline sheet-token construction is excluded.}
\label{tab:two_dataset_full_corpus}
\begin{adjustbox}{width=0.90\textwidth,center}
\begin{tabular}{llcccccc}
\toprule
\textbf{Dataset} & \textbf{Method} & \textbf{NDCG@5} &
\textbf{Macro R@5} & \textbf{Hit@5} & \textbf{MRR@5} &
\textbf{HN-FPR@1} $\downarrow$ & \textbf{Latency} $\downarrow$ \\
\midrule
IndustryTab-614 & LLM-only
& 0.5703$\pm$.0115 & 0.4194$\pm$.0288 & 0.8632$\pm$.0072
& 0.6983$\pm$.0130 & 0.0450$\pm$.0035 & 1.66\,s \\
& RAG
& 0.6344$\pm$.0042 & 0.4381$\pm$.0300 & 0.8667$\pm$.0105
& 0.7742$\pm$.0086 & 0.0277$\pm$.0183 & 8.97\,ms \\
& RAG+LLM
& 0.8374$\pm$.0174 & 0.6200$\pm$.0454 & 0.9644$\pm$.0053
& 0.9188$\pm$.0155 & 0.0196$\pm$.0140 & 2.37\,s \\
& SAT (Stage 1 only)
& 0.9072$\pm$.0188 & 0.6947$\pm$.0477 & 0.9736$\pm$.0121
& 0.9390$\pm$.0148 & 0.0092$\pm$.0020 & 8.79\,ms \\
& SAT
& \bestval{0.9173$\pm$.0102} & \bestval{0.7029$\pm$.0461}
& 0.9770$\pm$.0080 & \bestval{0.9492$\pm$.0038}
& \bestval{0.0069$\pm$.0060} & 15.13\,ms \\
\midrule
IndustryTab-1K & LLM-only
& 0.5248 & 0.4354 & 0.8547 & 0.6253 & 0.0355 & 2.37\,s \\
& RAG
& 0.6287 & 0.4722 & 0.8492 & 0.7705 & 0.0237 & 6.83\,ms \\
& RAG+LLM
& 0.8201 & 0.6526 & 0.9385 & 0.9201 & 0.0178 & 2.61\,s \\
& SAT (Stage 1 only)
& 0.8436$\pm$.0582 & 0.6771$\pm$.0626 & 0.9423$\pm$.0141
& 0.9069$\pm$.0258 & \bestval{0.0059$\pm$.0059} & 6.41\,ms \\
& SAT (Cross-Encoder)
& 0.8770$\pm$.0217 & 0.7213$\pm$.0164 & 0.9609$\pm$.0056
& 0.9142$\pm$.0149 & 0.0197$\pm$.0171 & 31.15\,ms \\
& SAT
& \bestval{0.9222$\pm$.0058} & \bestval{0.7590$\pm$.0051}
& \bestval{0.9683$\pm$.0065} & \bestval{0.9459$\pm$.0041}
& 0.0178$\pm$.0059 & 9.24\,ms \\
\bottomrule
\end{tabular}
\end{adjustbox}
\end{table*}

\begin{table*}[t]
\centering
\small
\setlength{\tabcolsep}{4.5pt}
\caption{BGE full-corpus sensitivity on IndustryTab-1K (seed 42).
Every row retrieves candidates from all 1{,}002 sheets and reports metrics on
the 179-query test split; no annotated positive is injected into the
candidate set. F+S denotes formula and summary-source dependency channels.
The reference is the seed-42 member of the three-seed SAT result in
\autoref{tab:two_dataset_full_corpus}. Higher is better except for HN-FPR@1.}
\label{tab:bge_sensitivity}
\begin{adjustbox}{width=0.90\textwidth,center}
\begin{tabular}{lllccccc}
\toprule
\textbf{Group} & \textbf{Setting} & \textbf{Candidate $K$}
& \textbf{NDCG@5} & \textbf{Macro R@5} & \textbf{Hit@5}
& \textbf{MRR@5} & \textbf{HN-FPR@1} \\
\midrule
Reference & F+S, 1 layer, gate $-4$ & 50
& \textbf{0.9155} & \textbf{0.7533} & 0.9609 & 0.9429 & 0.0118 \\
\midrule
Candidate pool & $K=10$ & 10
& 0.8577 & 0.6867 & 0.9553 & 0.9306 & \textbf{0.0000} \\
& $K=20$ & 20
& 0.8877 & 0.7194 & 0.9609 & 0.9385 & 0.0059 \\
\midrule
GNN depth & 2 layers & 50
& 0.9108 & 0.7505 & \textbf{0.9665} & 0.9412 & 0.0178 \\
& 3 layers & 50
& 0.9129 & 0.7473 & \textbf{0.9665} & 0.9437 & 0.0059 \\
\midrule
Gate init. & $-8$ & 50
& 0.9078 & 0.7454 & 0.9609 & 0.9429 & 0.0118 \\
& $-6$ & 50
& 0.8990 & 0.7364 & 0.9553 & 0.9320 & 0.0178 \\
\midrule
Dependency & Join & 50
& 0.9076 & 0.7430 & \textbf{0.9665} & \textbf{0.9453} & 0.0178 \\
channels & Aggregation & 50
& 0.9091 & 0.7444 & \textbf{0.9665} & 0.9401 & 0.0237 \\
& Formula & 50
& 0.9108 & 0.7485 & \textbf{0.9665} & 0.9415 & 0.0118 \\
& Summary source & 50
& 0.9069 & 0.7416 & \textbf{0.9665} & 0.9404 & 0.0296 \\
& Aggregation+Formula+Summary & 50
& 0.9111 & 0.7503 & \textbf{0.9665} & 0.9403 & 0.0178 \\
& All channels & 50
& 0.9093 & 0.7481 & 0.9553 & 0.9350 & 0.0178 \\
\bottomrule
\end{tabular}
\end{adjustbox}
\end{table*}

\paragraph{Metrics and candidate ceilings.}
NDCG@5 uses binary multi-label relevance; Macro Recall@5 is the mean fraction
of each query's relevant sheets captured in the top five; Hit@5 measures
whether at least one relevant sheet is returned; and MRR@5 uses the first
relevant rank. HN-FPR@1 is computed only over queries with annotated hard
negatives (288--289 per IndustryTab-614 split and 169 on IndustryTab-1K).
In the IndustryTab-1K seed-42 reference run, the fine-tuned BGE
top-50 candidate stage obtains 0.9467 Candidate Recall@50, 0.9944 Candidate
Hit@50, and contains all relevant sheets for 0.8659 of queries. Its
corresponding oracle Recall@5 is 0.7898.

The latency column in \autoref{tab:two_dataset_full_corpus} excludes model
loading and offline sheet-token construction because sheet representations
are cached after training. Stage 1 includes query tokenization and encoding,
similarity against all cached sheets, and top-$k$ selection ($k=5$ for
RAG and $k=50$ for RAG+LLM and SAT). SAT's 9.24\,ms total comprises 6.41\,ms
for retrieval and 2.82\,ms for candidate assembly and GNN reranking. LLM
latency includes prompt evaluation and constrained decoding. RAG and SAT
values are arithmetic means over 895 timed executions, while LLM values
average one complete pass over 179 queries.
The Cross-Encoder takes 31.15\,ms end to end, comprising 6.25\,ms for Stage~1
and 24.91\,ms for tokenizing and jointly encoding the 50 query--sheet pairs.
For IndustryTab-614, RAG and SAT average 870 timed executions (290 queries
and three repeats); LLM latency uses a separate sequential 30-query seed-42
timing pass, and RAG+LLM includes the 8.97\,ms retrieval stage.

\paragraph{Analysis.}
Qwen3.5-9B makes the LLM baselines substantially stronger than a small local
model. On IndustryTab-1K, LLM-only reaches 0.5248 NDCG@5, but processes
24,603 input tokens and takes 2.37\,s per query. RAG+LLM reduces the LLM
context by 8.7$\times$, to
2,830 tokens, and reaches 0.8201 NDCG@5, a 30.4\% relative improvement over
zero-shot RAG.
SAT Stage~1 adaptation raises NDCG@5 from 0.6344 to 0.9072 on IndustryTab-614
and from 0.6287 to 0.8436 on IndustryTab-1K. Training the GNN on the resulting
real top-50 candidate distribution reaches 0.9173 and 0.9222, respectively.
On IndustryTab-1K, the independently scoring Cross-Encoder reaches
0.8770 NDCG@5 at 31.15\,ms per query, while relational GNN refinement reaches
0.9222 with lower variation across seeds and $3.4\times$ faster inference.
On IndustryTab-614, SAT improves over RAG+LLM by 9.5\% relative
(0.9173 versus 0.8374).
On IndustryTab-1K, complete SAT exceeds the strongest RAG+LLM baseline by
12.5\% relative while reducing online latency from 2.61\,s to 9.24\,ms
($283\times$ faster).

\autoref{tab:two_dataset_full_corpus} therefore reports both retrieval quality
and deployable online latency without candidate injection.

\subsection{BGE Full-Corpus Sensitivity Analysis}
\label{sec:bge_sensitivity}

We conduct a strict full-corpus sensitivity analysis of the final BGE
pipeline. All variants in
\autoref{tab:bge_sensitivity} use the fine-tuned BGE Stage~1 retriever,
retrieve from all 1{,}002 IndustryTab-1K sheets, and evaluate the same 179
held-out queries with no positive-sheet injection. Stage~2 checkpoints are
selected using validation NDCG@5. The reference setting uses a top-50
candidate pool, one gated GNN layer, residual-gate initialization $-4$, and
formula plus summary-source dependency channels.
The reference row is the seed-42 run included in the three-seed SAT aggregate
reported in \autoref{tab:two_dataset_full_corpus}; consequently, its test
scores are not expected to equal the three-seed mean.

\begin{figure}[t]
\centering
\includegraphics[width=0.7\linewidth]{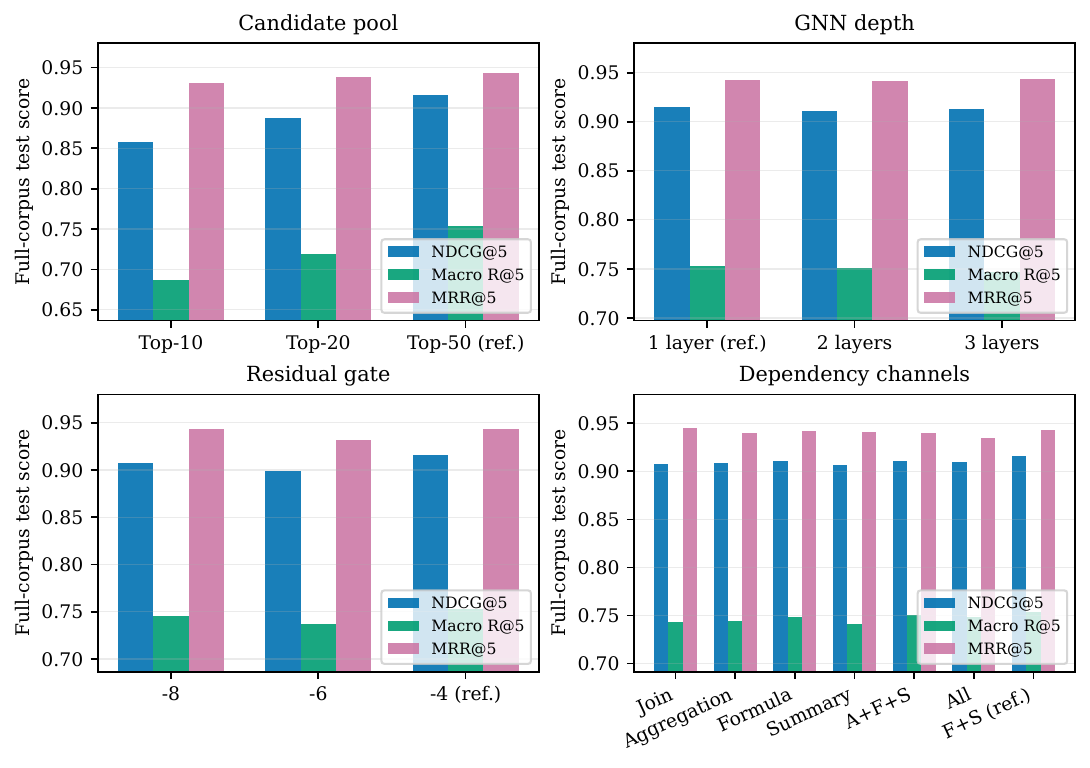}
\caption{BGE full-corpus sensitivity across candidate-pool size, GNN depth,
residual-gate initialization, and dependency-channel choice. The dashed line
marks the reference configuration. All values measure end-to-end retrieval
from the complete 1{,}002-sheet corpus.}
\label{fig:bge_sensitivity_metrics}
\end{figure}

\begin{figure}[t]
\centering
\includegraphics[width=0.7\linewidth]{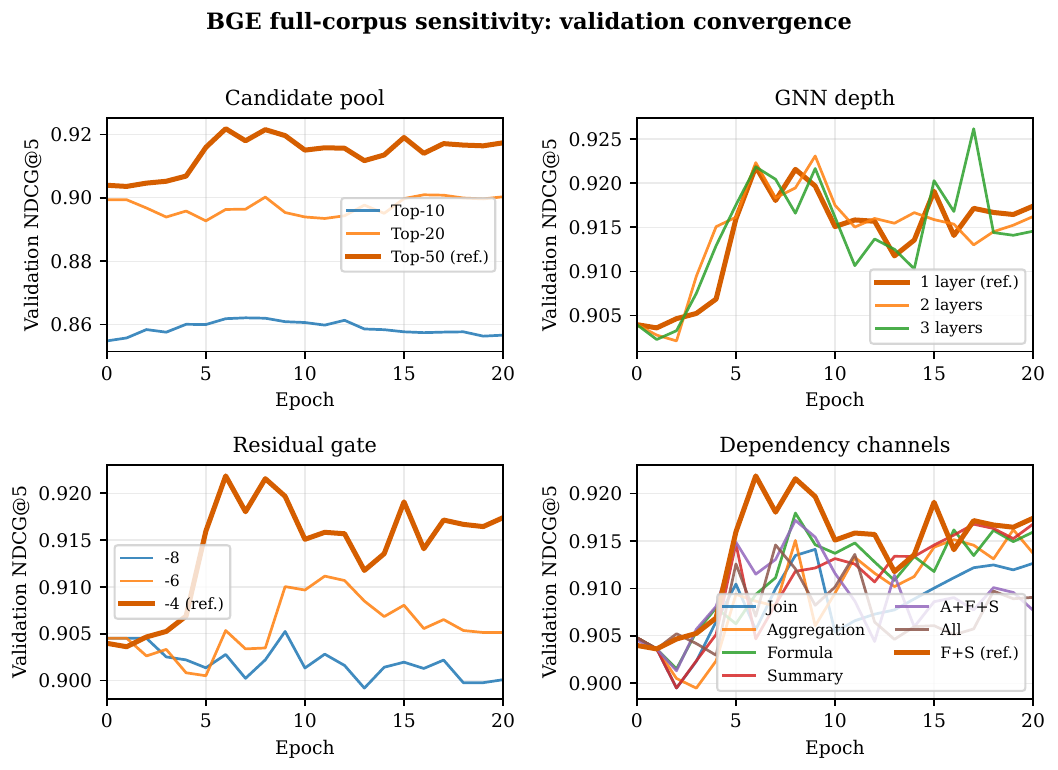}
\caption{Validation NDCG@5 during Stage~2 training for the BGE full-corpus
sensitivity runs. Candidate-pool size has the largest effect, while graph
depth and dependency-channel variants converge within a narrower range.}
\label{fig:bge_sensitivity_convergence}
\end{figure}

\autoref{fig:bge_sensitivity_metrics} summarizes the test-set effects of each
configuration choice, while \autoref{fig:bge_sensitivity_convergence} shows
the corresponding Stage~2 validation trajectories.

\paragraph{Candidate pool.}
Candidate recall remains the main end-to-end bottleneck. Increasing $K$ from
10 to 50 improves NDCG@5 by 0.0578 and macro Recall@5 by 0.0666 because the
reranker cannot recover relevant sheets omitted during candidate generation.

\paragraph{Graph configuration.}
The one-layer reference gives the highest test NDCG@5, while two and three
layers remain close and slightly improve Hit@5. A gate initialization of
$-4$ performs best; more negative initializations make graph updates too
conservative. Across dependency channels, formula plus summary-source edges
give the strongest NDCG@5. Join-only edges achieve the highest MRR@5 but lower
NDCG@5, and enabling all channels does not help, indicating that indiscriminate
edge addition introduces noise. Overall, the gated GNN is comparatively robust
once BGE supplies a sufficiently broad candidate pool.

\FloatBarrier

\section{Conclusion}
\label{sec:conclusion}
We introduced \textsc{Sheet As Token} (SAT), a two-stage retrieval framework
that represents each worksheet as a compact, cacheable semantic object and
uses a gated relational GNN to model explicit cross-sheet dependencies.
This design preserves sheet-level structure without full-workbook
serialization or fine-grained chunk fragmentation.

Across IndustryTab-614 and IndustryTab-1K, query--sheet BGE adaptation provides
the largest gain, while the complete GNN cascade reaches 0.9173 and 0.9222
NDCG@5, respectively. In the evaluated fixed-corpus setting, Stage~1 supplies
high-recall candidates and Stage~2 improves NDCG@5 and Macro Recall@5 without
positive injection. On IndustryTab-1K, this gain is accompanied by an
HN-FPR@1 increase from 0.0059 to 0.0178.

Our current evaluation does not capture all operational conditions of
production spreadsheet repositories, which evolve as workbooks are added,
removed, or revised. Future work will therefore study incremental Sheet Token
indexing, generalization to unseen workbooks, and retrieval efficiency in
production deployments.

\section*{Author Contributions}

The \textbf{Sheet Agent Research Team} consists of the following members: 
Yiming Lei (Georgia Institute of Technology),
Yuhang Yao (Carnegie Mellon University),
Yujia Zhang (University of Alberta),
Yiqi Wang (University of New South Wales),
Bo Guan (Northern Arizona University),
Depei Zhu (Effyic Intelligent Technology Co.),
Chunhui Wang (Peking University),
Zhuonan Hao (University of California, Los Angeles), and
Tianyu Shi (McGill University).

Yiming Lei, Yuhang Yao and Yujia Zhang contributed equally to this work. Chunhui Wang served as the corresponding author. Zhuonan Hao and Tianyu Shi co-led the project. All authors participated in the discussions, methodology design, and manuscript preparation. 

Corresponding emails: \texttt{yuhangyao8@gmail.com}, \texttt{znhao@g.ucla.edu}, \allowbreak \texttt{wangch2017@gsm.pku.edu.cn}, \allowbreak \texttt{tianyu.shi3@mcgill.ca}.

\bibliographystyle{unsrt}   
\bibliography{references}

\newpage
\appendix

\section{Detailed Gated-GNN Formulation}
\label{app:graph_details}

For each query $q$, Stage~1 retrieves a real candidate set
$\mathcal{C}(q)=\{S_1,\ldots,S_K\}$ from the complete corpus. Let
$\mathbf{z}_q$ and $\mathbf{z}_i$ be the normalized fine-tuned BGE
representations of the query and candidate sheet $S_i$. The base listwise
ranker initializes each node from their interaction:
\begin{equation}
 \mathbf{h}^{(0)}_i =
 \phi\!\left(
 [\mathbf{z}_q;\mathbf{z}_i;
  \mathbf{z}_q\odot\mathbf{z}_i;
  |\mathbf{z}_q-\mathbf{z}_i|]
 \right),
\end{equation}
where $\phi$ is an MLP and $\odot$ denotes element-wise multiplication.

\paragraph{Relational channels.}
The candidate graph contains semantic and query-conditioned similarities plus
available spreadsheet dependencies. Dependency edges represent formula
references, join-key compatibility, aggregation relations, and summary-source
links. Each enabled channel $r$ has a row-normalized adjacency matrix
$\widetilde{\mathbf{A}}_r$. A layer computes
\begin{equation}
 \mathbf{m}^{(\ell)}_{i,r}
 = \sum_j \widetilde{A}_{r,ij}
   \mathbf{W}^{(\ell)}_r\mathbf{h}^{(\ell)}_j,
 \qquad
 \Delta_i^{(\ell)}
 = \psi_\ell\!\left(
 [\mathbf{h}^{(\ell)}_i;
  \textstyle\sum_r \alpha_{i,r}^{(\ell)}
  \mathbf{m}^{(\ell)}_{i,r}]
 \right),
\end{equation}
where $\alpha_{i,r}^{(\ell)}$ is a learned query-conditioned relation weight.
The update is residual and gated:
\begin{equation}
 \mathbf{h}^{(\ell+1)}_i =
 \operatorname{LN}\!\left(
 \mathbf{h}^{(\ell)}_i+
 \sigma(g_\ell)\Delta_i^{(\ell)}
 \right).
\end{equation}
Initializing $g_\ell$ to a negative value makes the GNN begin close to the
validated MLP ranking and prevents untrained graph messages from destroying
the BGE ordering.

\paragraph{Ranking objective.}
The final node logit combines the base listwise score with a learned graph
correction. Training uses
\begin{equation}
 \mathcal{L} =
 \mathcal{L}_{\mathrm{listwise}}
 + \lambda_{\mathrm{node}}\mathcal{L}_{\mathrm{BCE}}
 + \lambda_{\mathrm{pair}}\mathcal{L}_{\mathrm{pair}},
\end{equation}
where the pairwise term separates every relevant sheet from retrieved hard
negatives in the same top-$K$ candidate set. Crucially, training and
evaluation use candidates retrieved by Stage~1 from the complete corpus; no
annotated positive is inserted.

\subsection{Training Hyperparameters}
\label{app:training_hyperparameters}

\autoref{tab:training_hyperparameters} reports the exact settings used for the
full-corpus results in Tables~\ref{tab:two_dataset_full_corpus}
and~\ref{tab:bge_sensitivity}. Sensitivity runs change only the variable named
by the corresponding row in Table~\ref{tab:bge_sensitivity}. For every seed,
we select the checkpoint with the highest validation NDCG@5.

\begin{table*}[t]
\centering
\small
\setlength{\tabcolsep}{4pt}
\caption{Training hyperparameters for the reported SAT pipeline. Stage~2 first
trains the listwise base scorer and then freezes it while training the gated
GNN. ``One epoch'' warmup denotes the number of optimizer steps in one
training pass.}
\label{tab:training_hyperparameters}
\begin{tabular}{p{0.18\textwidth}p{0.25\textwidth}p{0.25\textwidth}p{0.25\textwidth}}
\toprule
\textbf{Setting} & \textbf{Stage 1: BGE Sheet Encoder}
& \textbf{Stage 2: listwise base} & \textbf{Stage 2: gated GNN} \\
\midrule
Initialization
& BAAI/bge-base-en-v1.5; final four encoder layers and pooler trainable
& Fine-tuned 768-dimensional query and Sheet Tokens
& Selected listwise-base checkpoint; base scorer frozen \\
Training input
& Query, one relevant sheet, one retrieved hard negative, and in-batch negatives
& Real Stage~1 top-50 candidates
& The same top-50 candidates plus relation channels \\
Epochs / batch size
& 3 / 64 & 20 / 64 & 20 / 64 \\
Optimizer
& AdamW & AdamW & AdamW \\
Learning rate
& $2\times10^{-4}$ & $2\times10^{-4}$ & $5\times10^{-4}$ \\
Weight decay
& 0.01 & 0.01 & 0.01 \\
Schedule / warmup
& Linear / one-half epoch & Linear / one epoch & Linear / one epoch \\
Gradient clipping
& 1.0 & 1.0 & 1.0 \\
Maximum input length
& 256 tokens & Cached tokens & Cached tokens \\
Dropout / precision
& Backbone default / FP16 autocast & 0.1 / FP32 & 0.1 / FP32 \\
Objective
& Contrastive cross-entropy, $\tau=0.02$
& $\mathcal{L}_{\mathrm{listwise}}+0.5\mathcal{L}_{\mathrm{BCE}}
  +0.2\mathcal{L}_{\mathrm{pair}}$; margin 0.2
& Same ranking objective as the listwise base \\
Selection / seeds
& Validation NDCG@5 / 42, 43, 44
& Validation NDCG@5 / 42, 43, 44
& Validation NDCG@5 / 42, 43, 44 \\
\bottomrule
\end{tabular}
\end{table*}

The reported GNN uses one message-passing layer and keeps the base scorer
frozen. Four dense, row-normalized channels require no threshold: semantic
similarity, query relevance, schema Jaccard, and shape similarity.
IndustryTab-614 uses gate initialization $-6$ with aggregation, formula, and
summary-source dependencies. IndustryTab-1K uses initialization $-4$ with
formula and summary-source dependencies. Explicit dependency edges are binary
before row normalization.

The IndustryTab-1K Cross-Encoder runs use seeds 42--44 and independently score
each query--sheet pair among the Stage~1 top-50 candidates. They fine-tune
the last two BGE layers and a linear head for three epochs with batch size 64,
AdamW learning rate $2\times10^{-4}$, weight decay 0.01, maximum length 256,
FP16 autocast, and positive-class weight 4. Each query contributes up to two
positive and eight retrieved negative pairs. Validation selects NDCG@5 and the
blend weight between the Cross-Encoder logit and the Stage~1 score; the selected
weight is 10.

\section{Dataset Construction and Category Counts}

\label{app:data}

\subsection{Small Dataset: IndustryTab-614}
Table~\ref{tab:data_breakdown} gives a per-category breakdown of the
sheets contributing to the original corpus after Valentine-style pair
generation. The industrial block (top) is produced by our
template-based augmentor from six enterprise categories; the
personalized / public block (bottom) comes from four real-world
Chinese datasets that we translate and, where necessary, replace
high-cardinality free-text fields with typed placeholders.

\begin{table*}[!ht]
\centering
\caption{Per-category breakdown of IndustryTab-614. Some raw categories
(Engineering Quotes, Weibo Comments, Government Projects, Stock Forum
Sentiment) were dropped during preprocessing due to empty headers,
non-UTF-8 encodings, or row counts exceeding the fabricator's
input limit; these are documented in the release notes but do not
contribute sheets.}
\begin{tabular}{lll}
\toprule
Category & Source & \# Valentine pairs \\
\midrule
\multicolumn{3}{l}{\emph{Industrial (synthetic, template-augmented)}} \\
Financial Statements        & 18 simulated companies $\times$ multiple years & 126 \\
Sales Records               & quarterly, 2022--2024                          & 28  \\
Inventory Management        & inbound / outbound / ledger / stocktake        & 60  \\
Human Resources             & roster / payroll / attendance / training       & 24  \\
Project Management          & budget / progress / ledger                     & 18  \\
Stock Market Tick Data      & quarterly                                      & 24  \\
\midrule
\multicolumn{3}{l}{\emph{Personalized / public (real, translated)}} \\
Financial Business Data     & 18 institutional workbooks (Chinese)           & 18 \\
Global Stock Indices        & 10 international markets                       & 6  \\
Movie Review Tables (ZH)    & Chinese movie-rating tables                    & 2  \\
Text Sentiment Corpus       & sentence-level sentiment labels                & 1  \\
\midrule
\textbf{Total}              &                                                & \textbf{307 pairs/ 614 sheets} \\
\bottomrule
\end{tabular}
\label{tab:data_breakdown}
\end{table*}

\paragraph{Data sources.}
We construct IndustryTab-614 from a mixture of industrial and personalized real-world spreadsheets.
The industrial portion covers six enterprise categories (e.g., financial statements, human resources, project management), which we augment using a template-based generator to produce 210 synthetic workbooks spanning 18 simulated companies.
The personalized portion consists of four real-world public datasets, including institutional financial records and global stock indices.
All categorical cell values are translated to English, and high-cardinality free-text columns are replaced with typed placeholders (e.g., \texttt{Movie\_0001}, \texttt{Note\_0001}) to ensure reproducibility and mitigate PII leakage.
A full per-category breakdown is provided in Table~\ref{tab:data_breakdown}.

\subsection{Large Dataset: IndustryTab-1K}
\autoref{tab:data_breakdown_1k} reports the category composition of the
expanded corpus. It retains all 614 original sheet identities while increasing
temporal and entity coverage and adding Budget Planning. A metadata-level audit
finds 590 unchanged original records, 24 records with refreshed extracted
metadata, and 388 added sheet identities. All 501 Valentine source--target pairs
contribute two sheets, producing 1,002 sheets and 3,006 pairwise records (501
positive and 2,505 cross-table negatives). The listwise benchmark contains
1,797 queries with equally sized positive and negative sets.

\begin{table*}[t]
\centering
\small
\caption{Per-category sheet counts in IndustryTab-1K. Counts are measured
directly from the released \texttt{sheets.json}.}
\label{tab:data_breakdown_1k}
\begin{tabular}{lrlr}
\toprule
\textbf{Category} & \textbf{\# Sheets} &
\textbf{Category} & \textbf{\# Sheets} \\
\midrule
Financial Statements & 284 & Budget Planning & 172 \\
Synthetic Data & 140 & Inventory Products & 120 \\
Human Resources & 64 & Sales Data & 64 \\
Stock Market Data & 56 & Project Management & 48 \\
Financial Business Data & 36 & Global Stock Indices & 12 \\
Movie Ratings/Reviews & 4 & Text Sentiment Analysis & 2 \\
\midrule
\multicolumn{3}{r}{\textbf{Total}} & \textbf{1,002} \\
\bottomrule
\end{tabular}
\end{table*}

\paragraph{Dependency-aware query supervision.}
IndustryTab-1K contains 631 semantic-lookup, 357 comparison, 339 aggregation,
257 join-key, 140 formula-tracing, and 73 summary-source queries. In total,
1,319 queries are marked dependency-heavy. Each query also carries
schema-matched hard negatives (8.90 per query on average). The released
dependency graph contains 19,850 retained edges over 786 connected sheets:
17,806 join-key, 1,940 aggregation, and 104 formula-reference edges. The
unfiltered annotations additionally retain summary-source relations and
alternative aggregation/formula edges for controlled dependency ablations.

\paragraph{Pair generation.}
We use the Valentine fabricator~\cite{koutras2021valentine,valentinefabricator} with column overlap in $[0.5, 0.7]$ and row overlap of $\approx 0.5$.
Three noise channels are applied:
\begin{itemize}\itemsep=2pt\topsep=2pt\parskip=0pt
\item \textbf{Schema noise.} We use the table-name prefix rule from Valentine (e.g., \texttt{Vendor}$\to$\texttt{Sales\_Vendor}); the abbreviation and vowel-deletion rules are not applied.

\item \textbf{String noise.} Following Valentine's keyboard-proximity scheme, non-ASCII characters are first transliterated via \texttt{unidecode}; each resulting character on the source side is then independently replaced by a keyboard-proximity neighbor with probability $p=0.20$.

\item \textbf{Numerical noise.} For each numeric column on the source side with empirical mean $\hat{\mu}$ and standard deviation $\hat{\sigma}$, values are resampled from a perturbed half-normal:
\begin{equation}
\begin{aligned}
v'_i &\;\sim\; \bigl|\mathcal{N}(\mu',\,\sigma'^{2})\bigr|, \\
\mu' &= \hat{\mu}\,(1+\epsilon_\mu), \\
\sigma' &= \hat{\sigma}\,(1+\epsilon_\sigma), \notag
\end{aligned}
\label{eq:num_noise}
\end{equation}
where $\epsilon_\mu, \epsilon_\sigma \sim \mathrm{Uniform}([-0.5, -0.1] \cup [0.1, 0.5])$.
\end{itemize}

Each pair carries column-level alignment as supervision for pairwise sheet matching.

\paragraph{Translation and placeholders.}
Column names and low-cardinality categorical values are translated
via Google Translate with a 2{,}827-entry cache. High-cardinality
free-text columns are replaced with typed placeholders to keep the
released data small, reproducible, and free of PII:
\texttt{Movie\_NNNN} (movie titles, $\sim 22$k unique),
\texttt{Actor\_NNNN} (lead actors, $\sim 20$k),
\texttt{Director\_NNNN} ($\sim 9$k),
\texttt{Post\_NNNN} / \texttt{Author\_NNNN} (forum posts / authors,
up to $\sim 309$k unique), \texttt{User\_NNNN} (user handles), and
\texttt{Note\_NNNN} (free-text cells longer than 120 characters).
Strings below the 120-character cap are passed through the translator directly.

\end{document}